\pdfoutput=1

\documentclass[11pt]{article}

\usepackage[]{acl}

\usepackage{times}
\usepackage{latexsym}

\usepackage[T1]{fontenc}

\usepackage[utf8]{inputenc}

\usepackage{microtype}

\usepackage{inconsolata}

\usepackage{booktabs}
\usepackage{multirow}
\usepackage{comment}
\usepackage{amsmath}
\title{Constructing ASTE triples by searching phrases}
\title{Solving Aspect-Sentiment Triplet Extraction with search}
\title{ASTE-Transformer: Solving Aspect-Sentiment Triplet Extraction with search}
\title{ASTE-Transformer: Modelling Dependencies in Aspect-Sentiment Triplet Extraction}

\author{Iwo Naglik $^{1,3}$ \and
  Mateusz Lango $^{1,2}$\\
  $^1$ Poznan University of Technology, 
  Institute of Computing Science,
  Poznań, Poland \\
  $^2$ Charles University, Institute of Formal and Applied Linguistics, Prague, Czechia\\
  $^3$ deepsense.ai, (research@deepsense.ai) \\
  \texttt{inaglik.put.poznan@gmail.com}, 
\texttt{lango@ufal.mff.cuni.cz} }

\usepackage{graphicx}

\usepackage{tikz-dependency}
\begin{document}
\maketitle
\begin{abstract}
Aspect-Sentiment Triplet Extraction (ASTE) is a recently proposed task of aspect-based sentiment analysis that consists in extracting  (aspect phrase, opinion phrase, sentiment polarity) triples from a given sentence.
Recent state-of-the-art methods approach this task by first extracting all possible text spans from a given text, then filtering the potential aspect and opinion phrases with a classifier, and finally considering all their pairs with another classifier that additionally assigns sentiment polarity to them. 
Although several variations of the above scheme have been proposed, the common feature is that the final result is constructed by a sequence of independent classifier decisions.
This hinders the exploitation of dependencies between extracted phrases and prevents the use of knowledge about the interrelationships between classifier predictions to improve  performance.
In this paper, we propose a new ASTE approach consisting of three transformer-inspired layers, which enables the modelling of dependencies both between phrases and between the final classifier decisions.
Experimental results show that the method achieves higher performance in terms of F1 measure than other methods studied on popular benchmarks. 
In addition, we show that a simple pre-training technique further improves the performance of the model.

\end{abstract}

\section{Introduction}


Aspect-Sentiment Triplet Extraction (ASTE, \citealp{khw}) is a recent task in aspect-based sentiment analysis that involves the extraction of (aspect phrase, opinion phrase, sentiment polarity) triples from text. 
The aspect phrase denotes features or attributes of the described object, towards which the sentiment is expressed in the opinion phrase. 
The categorisation of this sentiment into typically three classes (positive/negative/neutral) is the final element of the triple.
For example, in the sentence "The hotel was very good" there is only one ASTE triple (hotel, very good, positive). See Fig~\ref{fig:example} for more examples.

\begin{figure}
\begin{dependency}[arc edge, arc angle=30]
\begin{deptext}[column sep=.1cm, font=\footnotesize]
The \& room \& was \& fine \& but \& the \& staff \& was \& rude.\\
\end{deptext}
\depedge{4}{2}{+}
\depedge{9}{7}{-}
\wordgroup[group style={fill=green!40, draw=brown, inner sep=.6ex}]{1}{2}{2}{a0}
\wordgroup[group style={fill=green!40, draw=brown, inner sep=.6ex}]{1}{7}{7}{a0}
\wordgroup[group style={fill=yellow!40, draw=brown, inner sep=.6ex}]{1}{4}{4}{a0}
\wordgroup[group style={fill=yellow!40, draw=brown, inner sep=.6ex}]{1}{9}{9}{a0}
\end{dependency}\\
Triplets: (room, fine, Positive), (staff, rude, Negative)\\
\begin{dependency}[arc edge, arc angle=30]
\begin{deptext}[column sep=.1cm, font=\footnotesize]
The  \& menu \&  is \&  limited \&  and \& extremely \&  pricy.\\
\end{deptext}
\depedge{4}{2}{-}
\depedge{6}{2}{-}
\wordgroup[group style={fill=green!40, draw=brown, inner sep=.6ex}]{1}{2}{2}{a0}
\wordgroup[group style={fill=yellow!40, draw=brown, inner sep=.6ex}]{1}{4}{4}{a0}
\wordgroup[group style={fill=yellow!40, draw=brown, inner sep=.6ex}]{1}{6}{7}{a0}
\end{dependency}\\
Triplets: (menu, limited, Negative), (menu, extremely pricy, Negative)
    \caption{Two examples of input sentences and ASTE triplets. The spans highlighted in yellow are opinion phrases, whereas spans highlighted in green are aspect phrases. The +/- sign denote positive/negative sentiment, respectively.}
    \label{fig:example}
\end{figure}

Since ASTE provides an answer to what? (aspect phrase), how? (sentiment), and why? (opinion phrase) questions regarding sentiment, it is sometimes referred to as a "near complete solution" to sentiment analysis~\cite{khw} and has attracted considerable research attention.
Several types of methods have been proposed, including sequence prediction~\citep{jet}, cascade processing~\citep{more-fine-grained}, prompting approaches~\cite{T5} or predicting a special word-by-word matrix~\cite{gts}.
However, the approaches that currently achieve the best predictive performance are span-level approaches~\cite{li-etal-2023-dual-channel,naglik}, which are the focus of this work.


Span-level methods~\cite{span-level} typically begin by extracting all possible text spans up to a predefined length from a given text. Each span  undergoes evaluation by a classifier to determine whether it contains an aspect phrase, an opinion phrase, or whether it should be excluded from further processing. 
The method then considers all possible pairs of identified opinion and aspect phrases, with a secondary classifier examining these pairs to discard false matches and assign sentiment polarity to the valid ones.



Although several modifications of this scheme have been proposed~\cite{sbc,simStar,liang2023stage}, they share the same property: the prediction of ASTE triples consists of dozens of independent classifier decisions, and the dependencies between decisions regarding the analysed spans are not modelled. 
This limits the predictive performance of these techniques, as some task-related knowledge simply cannot be learned.
Such unexploited properties of structured output include both deterministic rules and probabilistic patterns, some examples of which are given below (see more in App.~\ref{app:characteristics}). 
\begin{itemize}
    \item An opinion phrase assigned to multiple aspects, typically assign them the same sentiment polarity (see the 2$^{nd}$ example in Fig~\ref{fig:example}). 
    \item Two opinion phrases linked with a contrastive conjunction (like "but") and attached to one aspect phrase should have different sentiment polarities. A similar rule applies to opinions linked with correlative conjunctions ("and").
    \item Although one-to-many relations between aspect and opinion phrases are possible, the general probabilistic property is that constructing an increasing number of triples with a given phrase should be less and less likely.
    \item 
    An aspect phrase should only be extracted if it is associated with an opinion phrase. For instance, consider the word "room" in "The room was fine" and "I was given a single room". In the first sentence, this word should be extracted because it forms part of a triple (see Fig.~\ref{fig:example}), whereas in the second sentence, there is no associated opinion phrase.
\end{itemize}
Note that learning these properties requires joint modelling of the decisions to extract or link particular phrases. This is impossible to achieve in the current span-based ASTE frameworks, which often adapt end2end training but make a strong independence assumption and perform multiple independent classifier predictions.





In this paper, we address the challenge of modeling dependencies between the extracted phrases and between constructed triples by introducing ASTE-Transformer, a novel architecture for ASTE. 
Unlike conventional span-based ASTE approaches, our method does not perform multiple independent classifications to categorize extracted text spans into aspect/opinion/invalid phrases.  
Instead, aspect-opinion pairs are formed through a search in a specialized embedding space induced by modified self-attention mechanizm, where each span is represented twice: once as a potential aspect and once as a potential opinion phrase.
Furthermore, our approach does not construct final triples through independent classifications without considering other candidate triples; instead, for each candidate triple, it produces a representation that depends on all other triples.


The contributions of this paper are as follows:
\begin{itemize}
    \item We propose ASTE-Transformer, a new architecture composed of three types of transformer-inspired layers, that enables modelling the dependencies between the extracted phrases and between the constructed aspect-opinion pairs.
    \item To address the additional difficulty of training transformer models on relatively small ASTE datasets, we propose the simple idea of using pre-training on noisy supervised data that can be artificially generated from datasets for the more popular sentiment classification task.
    \item We carry out a fairly extensive experimental evaluation of the newly proposed method on four standard English benchmarks and two datasets for a more under-resourced language: Polish.  Ablation study and error analysis are also performed.
\end{itemize}

The experimental results demonstrated the superior predictive performance of ASTE-Transformer compared to other methods under study. 
Furthermore, the ablation study highlighted the importance of modelling dependencies, which accounted for improvements of up to 5 ppt on F1 score. 
Lastly, the proposed pre-training technique yielded additional, statistically significant performance improvements over previous state-of-the-art ASTE approaches.

\section{ASTE-Transformer}
The proposed method involves several processing steps, realised by three types of transformer-inspired layers: 1) standard transformer layers, 2) an aspect-opinion construction layer, and 3) a triple construction layer.

First, the input sentence is processed by a masked language model (MLM) composed of standard transformer layers that produce an embedding representation for each token.
Second, in line with span-based approaches, all text spans up to a certain length are extracted, and their corresponding embedding representations are constructed.
Next, the spans are analysed by an aspect-opinion pair construction layer, which searches for corresponding aspect-opinion phrases.
Finally, all candidate aspect-opinion pairs are processed by the triplet construction layer. This layer first computes a representation for each candidate pair, then a classifier assigns them a sentiment polarity or filters them out. 
Importantly, the constructed representation of an aspect-opinion pair depends on all the other candidate pairs.

\begin{figure*}
    \centering
    \includegraphics[width=\textwidth]{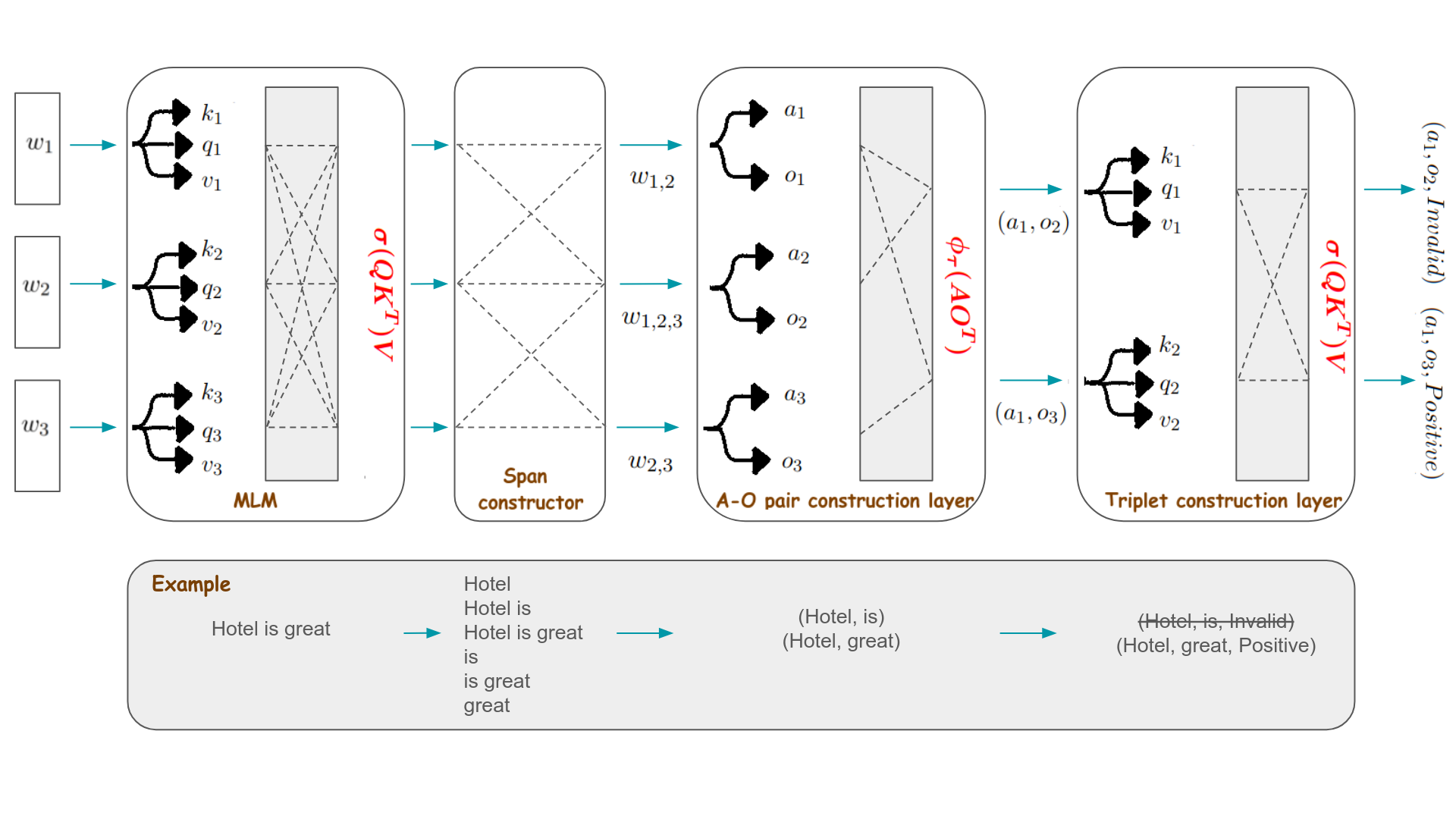}
    \caption{The overview of the proposed ASTE-Transformer architecture}
    \label{fig:architecture}
\end{figure*}

All the above-mentioned steps are realized by a single neural architecture, consisting of three types of transformer-like layers. The network is trained in an end-to-end fashion, by optimizing a loss function measuring the quality of constructed triplets and loss functions of additional intermediary tasks. 
An overview of the proposed neural network is shown in Fig.~\ref{fig:architecture}, and each of its parts is described in the following sections.

\paragraph{Problem formulation}
Given an input sentence $w_1, w_2, ..., w_n$, construct a set of ASTE triples $\{(a_i, o_i, y_i)\}_{i=1,..,m}$ where $a_i=\{w_j,w_{j+1},...,w_{j+|a_i|}\}$, $o_i=\{w_k,w_{k+1},...,w_{k+|o_i|}\}$ are aspect and opinion phrases consisting of one continuous text span, and $y_i\in \{Positive,Negative,Neutral\}$ is the polarity of sentiment expressed by $o_i$ towards the aspect mentioned in $a_i$.

\subsection{Contextualized representation}
\label{sec:mlm}
In the first part of our architecture, a distributed representation for each word in the input sentence is constructed by a transformer-based masked language model (MLM). 
Recall that in the transformer layer, a key $k_i$, value $v_i$ and query $q_i$ representations are constructed by fully-connected layers for each input word.
$$k_i = W_Kw_i \qquad v_i = W_Vw_i \qquad q_i = W_Qw_i$$
Then, according to the similarities between keys and queries (measured by the vector inner-product), a new word embedding $e_i$ representation is constructed by a weighted sum of value vectors.
$$e^{(w)}_i = \sigma(QK^T)_iV$$
where $\sigma$ is the softmax function, $K$ and $V$ are matrices containing $k_i$ and $v_i$ vectors for all input words, respectively.
Note, that the word representation $e_i$ is dependent on all the input words  $w_1, w_2, ..., w_n$.

\subsection{Span constructor}
\label{sec:spanconst}
In line with other span-based approaches, the next processing step is the extraction of all text spans up to a certain maximum length.
For instance, with the maximum length of 3, the following spans would be extracted: $\{w_1\}$, $\{w_1,w_2\}$, $\{w_1,w_2,w_3\}$, $\{w_2\}$, $\{w_2,w_3\}$, etc. 
The representation of each extracted span $s_i$ is constructed by max-pooling embeddings of words constituting it.
$$s_i = max\text{-}pooling(e_j, e_{j+1}, ..., e_k)$$
where $s_i$ is the representation of span starting from $j$-th word and ending at $k$-th word.

\subsection{Aspect-opinion pair construction layer}
\label{sec:pairlayer}
To match spans containing corresponding aspect-opinion phrases, we introduce a special transformer layer featuring a modified  the attention mechanism that performs pair matching.
This layer computes the distributed representations of each input span $s_i$ as potential aspect phrase $a_i$ and potential opinion phrase $o_i$ through fully-connected layers:
$$a_i = W_As_i \qquad o_i = W_Os_i$$
These representations are subsequently used to align opinions with their corresponding aspects through a search process, which involves computing similarities between aspect and opinion vectors and matching those with similarities exceeding a predefined threshold $\tau$.
$$p_i=[a_i;o_j] =  \phi_\tau(AO^T)$$
where $A$, $O$ are matrices containing vectors $a_i$, $o_i$ for all constructed spans and $\phi_\tau()$ is a thresholding operation, similar to attention masking, that from a given similarity matrix $S=AO^T$ extracts the indices $(i,j)$ of all the values $S_{i,j} > \tau$ above a threshold $\tau$.
The output of this layer is a set of extracted aspect-opinion pairs $p_i$, represented as a concatenation of aspect and opinion spans.

Note, that during the aspect-opinion pair construction, aspect and opinion phrases are considered jointly. 
This step lets us avoid initial categorization of spans into aspect and opinions phrases by a separate classifier applied multiple times. 

\subsection{Triplet construction layer}
\label{sec:tripletlayer}
During the final processing step, each extracted pair $p_i$ is either assigned a sentiment (positive, negative, or neutral) to create a triple or is dismissed as invalid. However, using a 4-class classification head for each pair $p_i$ on its own does not account for the interdependencies among the aspect-opinion triples while making predictions.
For example, as mentioned earlier, all triples with a given opinion phrase tend to have the same sentiment polarity. On the other hand, aspects with two opinions linked with contrasting conjunctions (like "but") will likely have opposite sentiments. 
It is not possible to model these and similar dependencies if the classification is performed completely independently.

Therefore, the extracted aspect-opinion pairs $p_i$ are processed jointly by an additional bidirectional transformer layer, which proved to be effective in modelling dependencies between the inputs for many tasks~\cite{devlin-etal-2019-bert}.
The input to this layer consists of representations of all extracted pairs $p_1, p_2, ..., p_N$ without typically added positional encoding, since the prediction should not vary on the arbitrary order of extracted pairs.
A classification head is then applied on top of the transformer layer with 4 classes: invalid, positive, negative, and neutral.
Predicting one of the last three classes results in the construction of a (aspect, opinion, sentiment) triple.


More formally, for each input aspect-opinion pair $p_i=[a_i,o_j]$, a new aspect-opinion embedding representation $e^{(p)}_i$ is constructed:
$$k_i = W_Kp_i \qquad v_i = W_Vp_i \qquad q_i = W_Qp_i$$
$$e^{(p)}_i = \sigma(QK^T)_iV$$
where $\sigma$ is the softmax function, $K$ and $V$ are matrices containing $k_i$ and $v_i$ vectors for all input aspect-opinion pairs, respectively. 
A softmax classifier then computes the final prediction using the constructed aspect-opinion embedding $e^{(p)}_i$:
$$y_i = \sigma(We^{(p)}_i)$$
Note that the aspect-opinion pair representation $e^{(p)}_i$ depends on all input pairs $p_1, p_2, ..., p_N$ returned by the aspect-opinion pair construction layer.

\section{Training procedure}
ASTE-Transformer model is trained via standard backpropagation in an end-to-end fashion. 
The training involves minimizing a composite loss function comprising the final ASTE loss, assessing the correctness of the constructed triples, along with two intermediate losses: the span selection loss and the aspect-opinion matching loss.
$$L = L_{ASTE} + L_{SpanSel} + L_{AO}$$
where $L_{ASTE}$ is the final ASTE loss, $L_{SpanSel}$ is the span selection loss,  and $L_{AO}$ is the aspect-opinion matching loss.

\paragraph{Span selection loss}
\label{sec:spanselloss}
To facilitate the matching of correct aspect and opinion phrases, we added an intermediary task of predicting whether the text span $s_i$ contains a valid aspect/opinion phrase. This is implemented as a simple binary task with valid/invalid outputs. 
Since the task suffers from heavy class imbalance, we applied Dice loss~\cite{li-etal-2020-dice} instead of standard cross-entropy:
$$ \hat{z}_i = \sigma(w^Ts_i +b) $$
$$L_{SpanSel}=\sum_{s_i \in Spans(w_{1..n})} \frac{2(1 - \hat{z}_i)^\alpha\hat{z_i}  z_i + \gamma}{(1 - \hat{z}_i)^\alpha\hat{z}_i + z_i + \gamma }$$
where $Spans(w_{1..n})$ generates all considered text spans, $\hat{z}_i$ is the estimated probability of the span $s_i$ being valid
, $w,b$ are additional weights
, $\alpha = 0.7$ is a scaling hyperparameter and $\gamma = 1$ 
is introduced for smoothing. 
The span representation $s_i$  is constructed in the span constructor (see Sec.~\ref{sec:spanconst}).

\paragraph{Aspect-opinion matching loss}
To promote the construction of a search space where correct aspect phrases are close to their corresponding opinion phrases, we apply a contrastive loss.
\begin{equation*} 
\begin{split}
L_{AO}=& \sum_{a_i \in A} \frac{\exp({a_i^To_{a_i}})}{
\sum_{o\in NegOpinions(a_i)} \exp({a_i^To})} \\
&+\sum_{o_i \in O} \frac{\exp({o_i^Ta_{o_i}})}{
\sum_{a\in NegAspects(o_i)} \exp({o_i^Ta})}
\end{split}
\end{equation*}
where $A$, $O$ are sets of all considered aspect and opinion phrases, $o_{a_i}$  is the representation of a correct opinion phrase for $a_i$, similarly  $a_{o_i}$  is the correct aspect phrase for $o_i$. 
The negative examples for a given aspect/opinion $NegOpinions(a_i)$ ($NegAspects(o_i)$) are constructed using hard mining, i.e.~four closest incorrect phrases are selected.

Since the aspect-opinion pair construction layer (Sec.~\ref{sec:pairlayer}) processes also incorrect aspect/opinion  phrases, we intend to push them away from all the phrases of opposite type. 
In this case, we also hard mine negative examples for the denominator of the loss function but in the nominator we put a constant instead of an inner-product with a corresponding correct phrase (as such does not exist). 


\paragraph{ASTE loss}
The construction of correct ASTE triples is enforced with the classification loss on the final $y_i$ with four possible outputs: positive, negative, neutral and invalid.
Due to the class imbalance of this task, the Focal loss~\cite{8237586} is applied instead of standard cross-entropy\footnote{Dice loss is a proper loss function only for binary classification task, so it cannot be used in this case.}.
$$L_{ASTE}=-(1-y_i)^{\gamma}\ln(y_i)$$
where $y_i$ is the probability of the correct class 
and $\gamma=2$ is a scaling hyperparameter. 
During training, all correct triples (even if not selected by previous layers) are passed to the final transformer classifier to fully utilize the learning information.

\section{Pretraining for ASTE}
Transformer-based architectures are known to benefit from previous pretraining, especially when dealing with limited supervised data.
In the proposed ASTE-Transformer, only the first part (MLM) is pre-trained using standard methods, while all subsequent layers are randomly initialized.
Therefore, we propose a simple idea of generating abundant noisy ASTE data from sentiment classification (SC) datasets and employing them for pretraining purposes.
Note that SC datasets are often much larger than ASTE datasets because they can be automatically collected from e-commerce platforms, where the consumer's overall product rating can be used as a proxy for opinion sentiment~\cite{ni-etal-2019-justifying}. 

The first step of our method is to train ASTE-Transformer model on the original ASTE dataset and apply it to texts from the SC dataset to produce artificial annotations. 
As predicting incorrect sentiment polarity is a factor negatively affecting the performance of ASTE models~\cite{yu-etal-2023-making}, we substitute the sentiment polarity in the generated triples with the gold standard sentiment of the whole sentence as provided in SC dataset. 
Finally, we train a new ASTE-Transformer from scratch, starting from pre-training it on the set of pseudo-labelled data, and then combining it with gold standard ASTE data. 
The last few training epochs are performed on gold standard ASTE data only.

\begin{table*}[t]
\centering\small
\begin{tabular}{l|rrr|rrr|rrr|rrr}
\toprule 
              & \multicolumn{3}{c}{14lap}           & \multicolumn{3}{c}{14res}                 & \multicolumn{3}{c}{15res}                 & \multicolumn{3}{c}{16res}           \\
              & Prec. & Rec. & F1             & Prec. & Rec. & F1                   & Prec. & Rec. & F1                   & Prec. & Rec. & F1             \\\hline
C-GPT 0-shot           &   n/a        &  n/a      &27.30         &    n/a       &     n/a   & 40.04               &   n/a        &     n/a   & 33.51              & n/a          &   n/a     & 42.18           \\
C-GPT 1-shot           &   n/a        &  n/a      &35.49        &    n/a       &     n/a   & 44.92              &   n/a        &     n/a   & 47.30              & n/a          &   n/a     & 50.09           \\
C-GPT 5-shot           &   n/a        &  n/a      &42.56        &    n/a       &     n/a   & 50.75             &   n/a        &     n/a   &49.99             & n/a          &   n/a     & 51.30          \\
Flan 1-shot           &   n/a        &  n/a      &5.19        &    n/a       &     n/a   & 9.26           &   n/a        &     n/a   &9.31             & n/a          &   n/a     & 11.81         \\\midrule

GAS           &   n/a        &  n/a      & 60.78          &    n/a       &     n/a   & 72.16                &   n/a        &     n/a   & 62.10                & n/a          &   n/a     & 70.10          \\
Pairing           &   n/a        &  n/a      & 61.68           &    n/a       &     n/a   & 72.53                &   n/a        &     n/a   &62.78                & n/a          &   n/a     & 71.38          \\
GTS           & 58.54     & 50.65  & 54.30          & 68.71     & 67.67  & 68.17                & 60.69     & 60.54  & 60.61                & 67.39     & 66.73  & 67.06          \\
PBF           & 56.60     & 55.10  & 55.80          & 69.30     & 69.00  & 69.20                & 55.80     & 61.50  & 58.50                & 61.20     & 72.70  & 66.50          \\
FTOP          & 57.84     & 59.33  & 58.58          & 63.59     & 73.44  & 68.16                & 54.53     & 63.30  & 58.59                & 63.57     & 71.98  & 67.52          \\
JET        & 55.39     & 47.33  & 51.04          & 70.56     & 55.94  & 62.40                & 64.45     & 51.96  & 57.53                & 70.42     & 58.37  & 63.83          \\
Span-ASTE     & 63.44     & 55.84  & 59.38          & 72.89     & 70.89  & 71.85                & 62.18     & 64.45  & 63.27                & 69.45     & 71.17  & 70.26          \\
SBC           & 63.64     & 61.80  & { 62.71}    & 77.09     & 70.99  & {73.92}       & 63.00     & 64.95  & 63.96                & 75.20     & 71.40  & {73.25} \\
SimSTAR&66.46 &58.23 &62.07 &76.23 &71.63 &73.86 &71.71& 59.59& 65.09& 72.02& 74.12& 73.06\\
STAGE-3D&71.98&53.86&61.58&78.58&69.58&73.76&73.63&57.90&64.79&76.67&70.12&73.24\\
{EPISA} & 66.98     & 60.55  & \underline{ 63.56} & 75.29     & 72.56  & { {73.89}} & 66.44     & 64.74  & {65.54}       & 71.12     & 72.45  & { 71.77}   \\

\midrule
Ours w/o pre.	&65.56&	60.36& {62.83} &74.51	&76.05&\underline{75.27}&	67.94&	67.91& \underline{67.89}&	74.96&	74.27  &\underline{74.61} \\
Ours w/ pre.	&	67.58	&62.48  &\textbf{64.90}&	76.43	&75.71  &\textbf{76.06}&	72.91	&71.34 & \textbf{72.10}&	76.27	&76.12&  \textbf{76.19} \\
\bottomrule
\end{tabular}
\caption{The experimental results of ASTE task on four English benchmark datasets. 
The best results according to F1-score are bolded, and the second-best results are underlined. C-GPT stands for Chat-GPT and Flan for Flan-UL2.}
\label{tab:res}
\end{table*} 

\section{Experimental evaluation}
\label{sec:experiments}
\subsection{Experimental setup}
\paragraph{Datasets} To evaluate the predictive performance of ASTE-Transformer, we conducted computational experiments on four ASTE datasets commonly used in related work: 14res, 14lap, 15res, 16res~\cite{khw,span-level,sbc}. 
Selected statistics of these datasets can be found in the Appendix~\ref{app:data-stats}.
Amazon Fine Food Reviews~\cite{mcauley2013amateurs} dataset was used for pretraining experiments, except for  experiments with 14lap, where Amazon Review Dataset (Digital Software)~\cite{ni-etal-2019-justifying} was used. For each domain, we utilize approx.~10,000 reviews.

\paragraph{Metrics} The performance of the models is measured with three metrics: precision, recall and F1-score. The extraction of an aspect/opinion phrase is considered correct only when it exactly matches the gold standard.
All reported metric values were computed on the corresponding test sets and averaged over four independent training runs.

\paragraph{Baselines} 
The method's results were compared with the results of GTS~\cite{gts}, PBF~\cite{more-fine-grained}, FTOP~\cite{First_Target_and_Opinion_then_Polarity}, GAS~\cite{T5}, 
JET~\cite{jet}, Span-ASTE~\cite{span-level}, SBC~\cite{sbc}, EPISA~\cite{naglik}, SimSTAR~\cite{simStar}, STAGE-3D~\cite{liang2023stage}, Pairing~\cite{yang2023pairing}.
All these methods are briefly described in Sec.~\ref{sec:rw}.
For reference, we also included the results obtained by~\citet{Zhang2023sentiment} with few-shot prompting of large language models (LLM): Chat-GPT and Flan-UL2.

\paragraph{Implementation} PyTorch implementation of our method and the code to reproduce experiments is publicly available\footnote{\url{https://github.com/NaIwo/ASTE-Transformer
}}.
Following related works, 
DeBERTa model~\cite{he2021deberta} was used as a MLM.
Similarly to other span-based approaches~\cite{span-level}, a pruning operation was applied to reduce the computational complexity (see App.~\ref{app:reducing}).

The model was optimized using Adam algorithm with default parameters.
The validation sets were used for early stopping and to select the threshold $\tau$ for the aspect-opinion pair construction layer.
All experiments were computed on one A100 GPU card.
Other implementation details are in App.~\ref{app:details}.

\begin{table}
    \centering\small
    \begin{tabular}{p{1cm}|p{1.5cm}llp{1.5cm}}
    \toprule
         &  14lap&14res&15res&16res\\\midrule

Ours w/o pre. &SBC, SimSTAR, STAGE-3D&None&None&SBC, SimSTAR, STAGE-3D\\\midrule
Ours w/~pre. &EPISA&None&None&None\\\bottomrule
    \end{tabular}
    \caption{Methods that yield a worse result on F1 score than the proposed method, but the difference is not statistically significant according to the T-test with significance level $\alpha=5\%$.
    }
    \label{tab:statsig}
\end{table}

\begin{table*}[]
\centering\small
\begin{tabular}{ll|rrr|rrr|rrr|rrr}
\toprule 
  &          & \multicolumn{3}{c}{14lap}           & \multicolumn{3}{c}{14res}                 & \multicolumn{3}{c}{15res}                 & \multicolumn{3}{c}{16res}           \\
   Pre.      & Trans.      & Prec. & Rec. & F1             & Prec. & Rec. & F1                   & Prec. & Rec. & F1                   & Prec. & Rec. & F1             \\
\midrule
No&No	&
64.93	&59.12 &61.85&
 	73.67&	75.61 &74.62&
	65.41&	69.69& 67.45&
	72.43&	73.20& 72.78\\

No&Yes	&65.56&	60.36& \textbf{62.83} &74.51	&76.05&\textbf{75.27}&	67.94&	67.91& \textbf{67.89}&	74.96&	74.27  &\textbf{74.61} \\\midrule
Yes&No	&	
	66.42&	63.12&64.73&
	74.22&	76.29&75.23&
	68.50&	69.18& 68.83&
	74.51&	74.90&74.70
\\
Yes&Yes	&	67.58	&62.48  &\textbf{64.90}&	76.43	&75.71  &\textbf{76.06}&	72.91	&71.34 & \textbf{72.10}&	76.27	&76.12&  \textbf{76.19} \\
\bottomrule
\end{tabular}
\caption{The results of an ablation study of the proposed method with/without pretraining (Pre.) and with/without transformer layer for triplet construction (Trans.) that models dependencies between candidate triples. }
\label{t:res-abla}
\end{table*} 
\subsection{Evaluation of model performance}
The main experimental results are presented in Table~\ref{tab:res} and a brief summary of performed statistical tests is presented in Table~\ref{tab:statsig}.

Comparing the methods using the same training data (i.e. ~without pre-training), ASTE-Transformer achieves the highest F1 score on three out of four benchmark datasets. On the remaining dataset (14 laps), it is the second-best method, surpassed only by EPISA. The high performance of the method seems to be the result of improving recall without significantly degrading precision.

The use of our simple pre-training method further improved the performance of ASTE-Transformer, resulting in the highest F1 score for all datasets.
The difference between ASTE-Transformer and all other methods is statistically significant for three datasets.
For the remaining 14-lap dataset, the difference is not statistically significant only when compared to EPISA.

We also investigated the usefulness of the proposed pretraining for one additional method, EPISA. The results are presented in Tab.~\ref{tab:res-pretraining}. For three out of four datasets, the use of our simple pretraining procedure was also beneficial for EPISA, with the highest improvement on the 15res dataset of almost 2 ppt. 
In general, however, the improvements are much smaller than for ASTE-Transformer. This may indicate that the EPISA model has a smaller capacity compared to the ASTE-Transformer and therefore cannot fully benefit from additional pre-training.
\begin{table}
\centering\small
\begin{tabular}{l|rrrr}
\toprule 
Method&14lap&14res&15res&16res\\\midrule
EPISA w/o pre. & \underline{ 63.56} & {73.89} &  {65.54}       &  { 71.77}   \\
EPISA w/ pre. &62.77&74.07&67.87&72.22\\
\midrule
Ours w/o pre.	& {62.83} &\underline{75.27}&	\underline{67.89}  &\underline{74.61} \\
Ours w/ pre.	&	\textbf{64.90}&	\textbf{76.06}&	 \textbf{72.10}& \textbf{76.19} \\
\bottomrule
\end{tabular}
\caption{The experimental results of EPISA and ASTE-Transformer with and without pretraining. }
\label{tab:res-pretraining}
\end{table} 
\subsection{Ablation study}
To verify the effectiveness of using a triplet representation that takes into account the dependencies between all candidate triples, we performed an ablation study where our triplet construction layer was replaced with a standard fully-connected layer.
 Both the results with and without pretraining are reported in Table~\ref{t:res-abla}.

 In both scenarios, i.e. with and without pretraining, the version of the ASTE-Transformer  with the triplet construction layer achieved better results than the fully connected layer, offering improvements of up to 3 ppt on F1 score.
 The ablation study also confirms the effectiveness of our pretraining technique, since for both variants of the ASTE-Transformer architecture, pretraining improves the results on all datasets (up to 4 ppt on F1).


\subsection{Evaluation on other languages}

In contrast to most related work, which only runs experiments on English, we also ran evaluations on two recent ASTE datasets for Polish~\cite{lango-etal-2024-polish}. In all methods, MLM was replaced by Polish TrelBERT~\cite{szmyd-etal-2023-trelbert}.

The results presented in Table~\ref{tab:polish} show that the ASTE Transformer obtained the highest F1 score on both datasets.
As the texts in the Polish datasets contain on average more triples and a higher number of more difficult one-to-many relations (see App.~\ref{app:data-stats}), we also report the results of our method without the final transformer layer modelling dependencies. As expected, we observe even more significant improvements compared to the ablation experiment on English.

\begin{table}[]
    \centering\small
    \begin{tabular}{cc|ccc}
    \toprule
    Dataset&Method&Prec&Rec&F1\\
    \midrule
       \multirow{4}{*}{products}&GTS  &  45.15& 40.17 &41.74\\
       &EPISA  &  50.01 & 43.36 &46.07\\\cmidrule(l){2-5}
       &Ours  w/o trans. &43.07 &	43.76 &	43.20\\
       &Ours &  46.87& 47.57 &\textbf{46.89}\\
       \midrule
         \multirow{4}{*}{hotels}&GTS  &  42.07 &37.82& 39.08\\
       &EPISA  & 49.07 &41.66& 44.72\\\cmidrule(l){2-5}
       &Ours w/o trans. &45.81&35.70&	39.91\\
       &Ours   &  47.79& 42.35 &\textbf{44.75}\\
       \bottomrule
    \end{tabular}
    \caption{The experimental results of ASTE task on two Polish benchmark datasets. Additionally, we report ablation of our method without final transformer layer.}
    \label{tab:polish}
\end{table}

\subsection{Visualization of the search space}
\begin{figure}[t]
    \centering
    \includegraphics[width=\columnwidth]{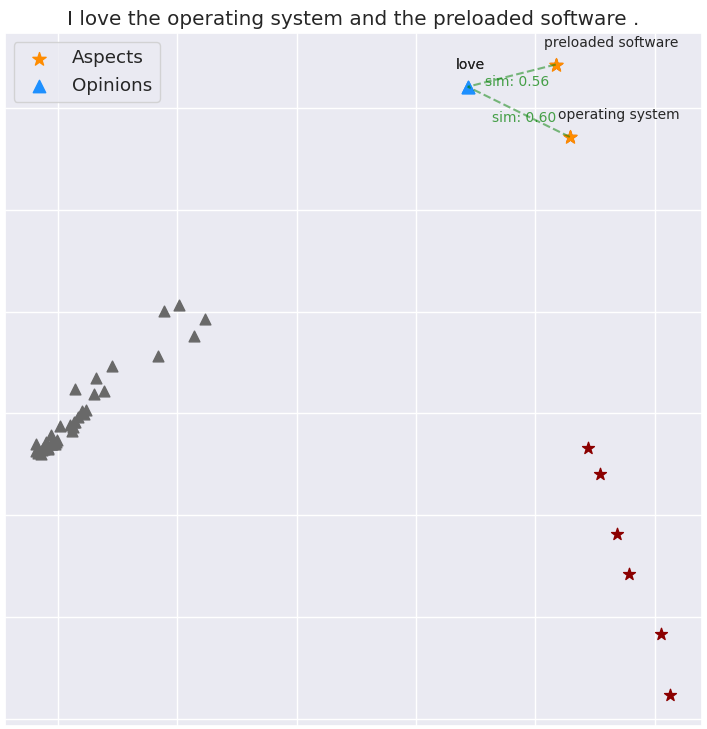}
    \caption{Visualization of the search space in the aspect-opinion pair construction layer for the sentence: "I love the operating system and the preloaded software". The gold standard triples are (operating system, love, Positive), (preloaded software, love, Positive).
    }
    \label{fig:viz1}
\end{figure}

To better understand how the search mechanism works in the proposed aspect-opinion pair construction layer, we visualized the induced embedding space using PCA for randomly selected test instances. An example of such a visualization is shown in Fig.~\ref{fig:viz1} and further examples can be found in App.~\ref{app:viz}.
In most of the visualizations, we observe that the representations of correct aspect and opinion phrases are close to each other. Moreover, the groups of invalid aspect and opinion phrases are clearly separated, placed at a considerable distance.

\subsection{Error analysis}
We investigate how specific components of our model affect the outcome by measuring several intermediate metrics: 1) the performance of an additional binary classifier trained to determine the validity of a span based on its representation $s_i$ (Span binary), 2) the effectiveness of extracting correct aspect-opinion pairs in the aspect-opinion pair construction layer (A-O pair layer), 3) the classification performance of the final sentiment classifier for assigning sentiment to triples or filtering them (Final 4-class), and 4) the classification performance of the same classifier when tasked with recognizing valid/invalid triples in a binary manner (Final binary). The results are showcased in Tab.~\ref{tab:error} and in App.~\ref{app:error}.

We observe that the span representation $s_i$ computed in the span constructor layer already encodes the information whether the span is a valid aspect or opinion phrase, as the simple binary classifier was able to distinguish them with 84\% F1 score.
The pairs constructed by the aspect-opinion pair construction layer have high recall and only slightly lower precision.  
The final classifier has a high performance in discriminating between valid and invalid triples, but assigning sentiment polarity to the triples seems to be more challenging.

\begin{table}[]
\small \centering
\begin{tabular}{l|lll}
\toprule
               & Prec.   & Rec.   & F1     \\\midrule
Span Binary    & 84.10   & 85.62  & 84.84  \\
A-O pair layer & 68.68   & 72.78  & 66.34  \\
Final 4-class  & 69.53   & 69.53  & 69.53  \\
Final binary   & 97.21   & 76.00  & 85.28  \\
Overall        & 67.58   & 62.48  & 64.90 \\
\bottomrule
\end{tabular}
\caption{The performance of several parts of ASTE-Transformer measured on 14lap dataset.}
\label{tab:error}
\end{table}

\section{Related works}
\label{sec:rw}
Since the introduction of Aspect Sentiment Triplet Extraction (ASTE) by~\citet{khw}, various approaches were proposed for this task.

JET~\cite{jet} converts the problem into a sequence labelling task using enhanced BIOES tagging schema.
Similarly, PBF~\cite{more-fine-grained} uses three sequential tagging predictors to construct triplets. 
A different approach is to encode ASTE triples in a word-by-word matrix with Grid Tagging Scheme (GTS,~\citealp{gts}).
Each element of this matrix is predicted by an independent classifier.

In contrast to GTS which relies on word-to-word interactions, Span-ASTE~\cite{span-level} considers all possible text spans from the input sentence, performing multiple independent classifications to construct the output.
SBC~\cite{sbc} uses span representations constructed with a special separation loss and has bidirectional structure to generate aspect-opinion pairs.
FTOP~\cite{First_Target_and_Opinion_then_Polarity} divides the input sentence into opinion/aspect phrases using sequence prediction and considers all aspect-opinion pairings by a classifier. 
Recently, EPISA~\cite{naglik} explored the possibility of making the predictions dependent while constructing ASTE triples. In contrast to our method, EPISA uses a 2-dimensional CRF over a decision matrix, which is intractable without making additional assumptions about pairwise decision independence and approximating potential functions with Gaussian kernels.

Another group of approaches combines span-based and matrix prediction approaches by predicting a matrix with span-based tags. Such approaches include STAGE~\cite{liang2023stage} and SimSTAR~\cite{simStar}. 
Both these approaches predict matrices by applying softmax classifiers independently.
Finally, \citet{T5} presented a generative approach called GAS, which uses prompting of the T5 language model.
Some of the generative methods also use contrastive learning to learn better representations~\cite{yang2023pairing}, but they do not use search to pair aspects with opinions.

A data augmentation technique for ASTE was proposed in~\cite{zhang-etal-2023-target-source}, but in comparison to our simple pretraining idea, it is rather complex as it employs reinforcement learning and trains additional generator and discriminator models.

\section{Summary}

In this paper, we have demonstrated the potential of exploiting dependencies between constructed triples in span-based ASTE approaches. The proposed ASTE-Transformer method showed superior predictive performance on both English and Polish benchmarks. 
Additionally, a simple pre-training scheme proved to further improve the performance.


\section*{Acknowledgements}
This research has received funding from the National Center for Research and Development under the Infostrateg program (project: INFOSTRATEG-III/0003/2021-00 “Development of an IT system using AI to identify consumer opinions on product safety and quality” realized in a consortium of Poznan Institute of Technology and Poznan University of Technology).

\section*{Limitations}
This work addresses the issue of performing multiple independent classifications by span-based ASTE approaches to produce the final result.
This is achieved by introducing an aspect-opinion matching layer and constructing interdependent triplet representations. Although this promotes the exchange of information about the considered triples, the classifier predictions on top of this mutually dependent representation are still independent in a probabilistic sense. To make the decisions dependent, some methods such as CRF have been proposed for sequences and graphs, but we are not aware of similar methods for sets, which is the case considered in this paper. 
Note that it has been shown that the use of interdependent representations is an effective way to explore information about dependencies for sequence prediction, as it significantly reduces the possible performance gains from making the classifier's predictions strictly dependent~\cite{reimers-gurevych-2017-reporting}.

Additionally, this work uses pre-trained language models, which are known to expose certain social biases reflected in their training data.

\bibliography{anthology,custom}

\appendix

\section{Dataset details}
\label{app:data-stats}
The basic characteristics of benchmark datasets are given in Table~\ref{tab:data-stats}.
\newcommand*{\tabindent}{ \hspace{3mm}}
\begin{table*}[]
\centering\small 
\begin{tabular}{p{7cm}|rrrr|rr}
\toprule
& \multicolumn{4}{c|}{English}   & \multicolumn{2}{c}{Polish}             \\
& 14lap & 14res & 15res & 16res & hotels     & products               \\\midrule
Number of sentences                                                                      & 1453 & 2068 & 1075 & 1393 & 590  & 511            \\\midrule
Number of triplets                                                                       & 2349 & 3909 & 1747 & 2247 & 1197 & 851   \\
\tabindent incl. with negative sentiment                                                        & 774  & 754  & 401  & 483  & 541  & 376   \\
\tabindent incl. with neutral sentiment                                                         & 225  & 286  & 61   & 90   & 58   & 54    \\    
\tabindent incl. with positive sentiment                                                          & 1350 & 2869 & 1285 & 1674 & 598  & 421   \\ \midrule
Number of aspect phrases                                                             & 2030 & 3392 & 1507 & 1946 & 798  & 693    \\
\tabindent incl. single word aspect phrases                                         & 1292 & 2545 & 1102 & 1427 & 681  & 526    \\
\tabindent incl. multi-word aspect phrases                                           & 738  & 847  & 405  & 519  & 117  & 167    \\\midrule
Number of opinion phrases                                                            & 2030 & 3409 & 1620 & 2078 & 1156 & 827    \\
\tabindent incl. single word opinion phrases                                              & 1705 & 3037 & 1421 & 1829 & 412  & 343    \\
\tabindent incl. multi-word opinion phrases                                           & 325  & 372  & 199  & 249  & 744  & 484  \\\midrule
Number of one-to-many relation  & 535  & 812  & 307  & 388  & 323  & 150    \\
\tabindent incl. one aspect-to-many opinions                                           & 281  & 443  & 208  & 263  & 289  & 128   \\
\tabindent incl. one opinion-to-many aspects                                           & 254  & 369  & 99   & 125  & 34   & 22  \\\midrule
Number of triplets w/ single words spans                                   & 1305 & 2631 & 1140 & 1478 & 385  & 287   \\
Number of triplets w/ multi-word phrases                       & 1044 & 1278 & 607  & 769  & 812  & 564   \\
\tabindent incl.~with multi-word opinion and single-word aspect  & 207  & 302  & 149  & 188  & 649  & 377   \\
\tabindent incl.~with multi-word aspect and single-word opinion  & 684  & 875  & 403  & 513  & 46   & 70  \\\midrule
Mean sentence length (words)                                                    & 18.4 & 16.9 & 15.0 & 14.9 & 16.4 & 21.0 \\
Mean length of aspect phrases                        & 1.47 & 1.40 & 1.45 & 1.44 & 1.26 & 1.40  \\
Mean length of opinion phrases                    & 1.25 & 1.16 & 1.19 & 1.19 & 2.97 & 2.22           \\
\bottomrule
\end{tabular}
\caption{Selected quantitative characteristics of benchmark datasets}
  \label{tab:data-stats}
\end{table*}

\section{Implementation details}
\label{app:details}
The structure of all linear layers adheres to a consistent architectural block pattern: LayerNorm, followed by a Linear layer, a ReLU activation function, and a Dropout layer with a rate of 0.1.

For the purpose of span selection loss (see Sec.~\ref{sec:spanselloss}), the classifier consisted of four linear blocks. The first layer had an input dimension of 768, which corresponds to the dimensionality of MLM embedding. Subsequent layers followed a halving dimension strategy: 768/2, 768/4, and 768/8, culminating in a layer that flattens the output to a single logit for binary classification.

The implementation of the pair constructor layer (Sec.~\ref{sec:pairlayer}) also integrates linear blocks. The dimensionality progression for these layers is as follows: starting from an initial dimension of 768, it moves through subsequent dimensions of 768/2, 768/4, 768/2 and the result from these layers is used as a representation of an aspect or opinion. These representations are then used to match corresponding opinions to their aspects by performing a search i.e. computing similarities between aspect and opinion vectors.

The final transformer-based classifier (Sec.~\ref{sec:tripletlayer}) utilized the TransformerEncoder\footnote{\url{https://pytorch.org/docs/stable/generated/torch.nn.TransformerEncoder.html}} class from the PyTorch library. The input is constructed by the concatenation of the embeddings for aspect/opinion spans, CLS token and an embedding representing the distance between aspect/opinion spans. Next, this concatenation result is passed through the linear layer to get, a 4 times smaller, 584-sized dimension and this value is an input to a transformer layer. The transformer's attention mechanism has 4 attention heads, ensuring a multi-head perspective in processing the input data. The following linear layer reduces dimensionality to 4 logits which are used to make the final prediction regarding a given pair of spans.
During training, all correct phrases, even those below $\tau$, are passed to the final transformer classifier to fully utilize the learning information.

For training the model, PyTorch Lightning\footnote{\url{https://lightning.ai/docs/pytorch/stable/common/trainer.html
}} library was used. The training was scheduled to last for a minimum of 30 epochs and a maximum of 130 epochs, incorporating gradient clipping set at 0.8 to mitigate the risk of exploding gradients. 

During the test phase, the threshold $\tau$ for span filtering was set slightly higher than during training (see App.~\ref{app:tau}). This adjustment allowed more phrase pairs to pass through the pair construction layer during the training phase, to enhance the model's capability to reject irrelevant spans at later stages. However, in testing, the trained model exhibits improved embedding quality, justifying a higher threshold for span filtering to reduce the risk of false positives. It is noteworthy that the decision about $\tau$ being used was made based on precision and recall curves calculated on the validation set, as in Fig. \ref{fig:tau} (see App.~\ref{app:tau} for details).

A simple filtering heuristic was employed to refine the model's output further. This involved removing overlapping spans from the output and retaining those with higher probabilities in cases of conflict. Such a strategy enhanced the precision of the model by prioritizing the selection of the most probable span predictions, thus contributing to the overall efficacy and reliability of the ASTE-Transformer.



\section{Reducing the computational complexity of ASTE-Transformer}
\label{app:reducing}
Since span-based approaches analyse all text spans up to a certain length, different techniques are used to reduce their computational complexity. 
Some span-based approaches~\cite{span-level} use a pruning operation that takes into account the results of the valid/invalid span classification. Since we aim to improve model performance by exploiting the dependencies between model predictions, a filtering approach based on multiple independent classifications was not a viable option.

Inspired by EPISA~\cite{naglik}, we train a CRF tagger on MLM representations to split the input sentence into spans by BIO-tagging both opinion and aspect phrases. The output of a tagger is augmented by producing all spans that are up to 1 word longer in each direction (i.e. starting at an earlier position or ending at a later position than indicated by the phrase boundaries predicted by the tagger).
We found that this technique is very effective in extracting aspect phrases
, but fails to extract opinion phrases with sufficient quality. 
Therefore, in the aspect-opinion matching layer (see Section~\ref{sec:pairlayer}), an opinion representation $o_i$ is computed for each possible text span, but the aspect representation $a_i$ is computed only for the spans contained in the tagger result (with the aforementioned augmentation). 
This was sufficient to reduce both the memory and computational requirements of our approach. 
On one A100 GPU card, training on the datasets considered in the paper typically takes about 80 minutes.
To make the experimentation easier, we train the CRF tagger jointly with ASTE-Transformer.

The computational complexity of the aspect-opinion matching layer could be reduced in many other ways, for example by replacing the naive implementation of searching for matching phrases with more sophisticated Maximum Inner Product Search (MIPS) techniques. These include fast approximate search techniques, which have already been shown to make transformer architectures faster without compromising output quality~\cite{kitaev2020reformer},

\section{Additional visualizations of the aspect-opinion search space}
\label{app:viz}
\begin{figure*}
    \centering
    \includegraphics[width=.7\textwidth]{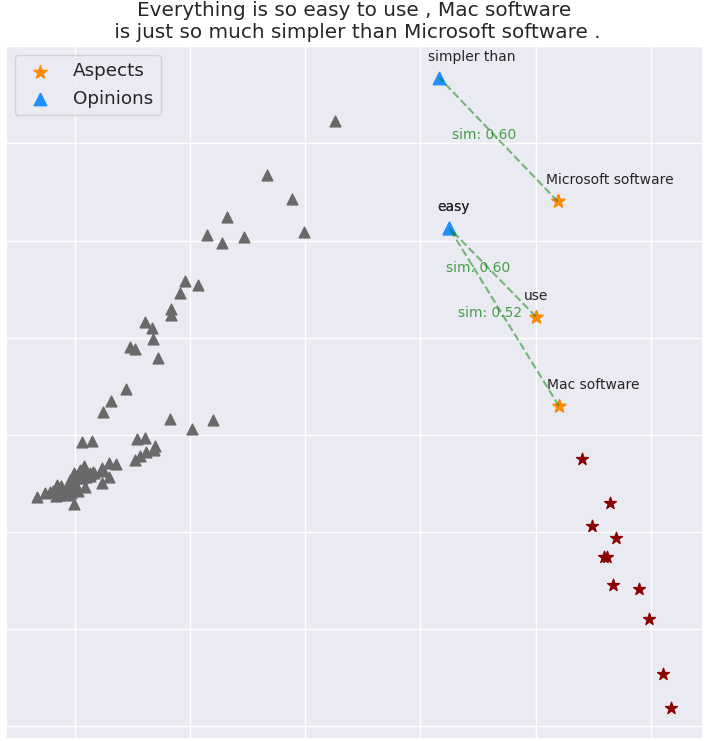}
    \caption{Visualization of search space in aspect-opinion pair construction layer for the sentence: "Everything is so easy to use, Mac software is just so much simpler than Microsoft software.".}
    \label{fig:viz2}
\end{figure*}

\begin{figure*}
    \centering
    \includegraphics[width=.7\textwidth]{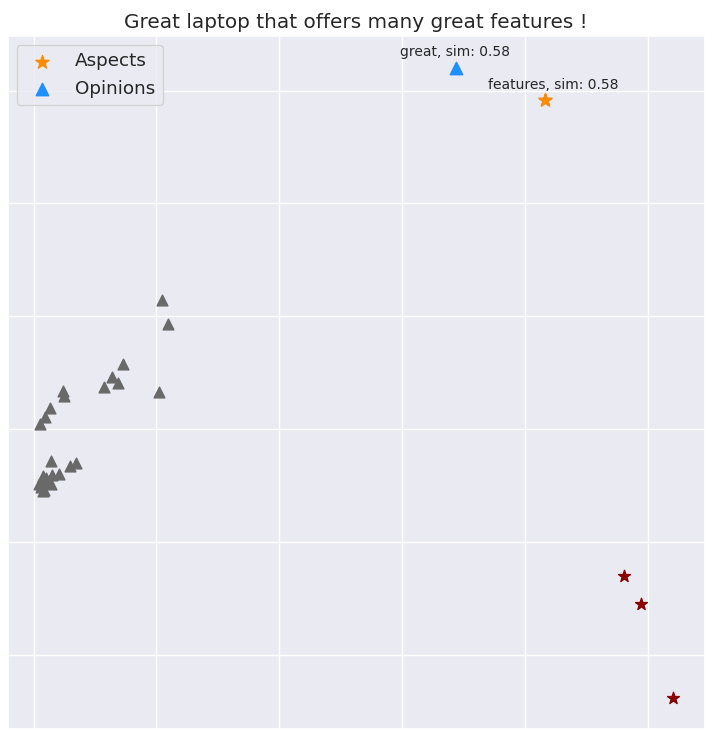}
    \caption{Visualization of search space in aspect-opinion pair construction layer for the sentence: " Great laptop that offers many great features!".}
    \label{fig:viz3}
\end{figure*}

The visualization of the search space produced by aspect-opinion matching layer for two additional example sentences is provided in Fig.~\ref{fig:viz2} and Fig.~\ref{fig:viz3}.

\section{The impact of $\tau$ threshold}
\label{app:tau}
The value of $\tau$ threshold used in the aspect-opinion pair matching layer influences the final result by controlling how many aspect-opinion pairs will be forwarded to further layers.
Since the classifier at the end of ASTE-Transformer has the possibility of filtering incorrect pairs by assigning them an "invalid" class, producing superfluous pairs at this stage of processing is not very detrimental.
However, the lack of construction of a correct aspect-opinion pair has a direct negative influence on the result, as such a pair cannot be constructed by other layers.
On the other hand, producing too many pairs negatively influences the processing time of further layers and can compromise the predictive performance by adding too many noisy pairs for further processing.

\begin{figure*}
    \centering
    \includegraphics[width=\textwidth]{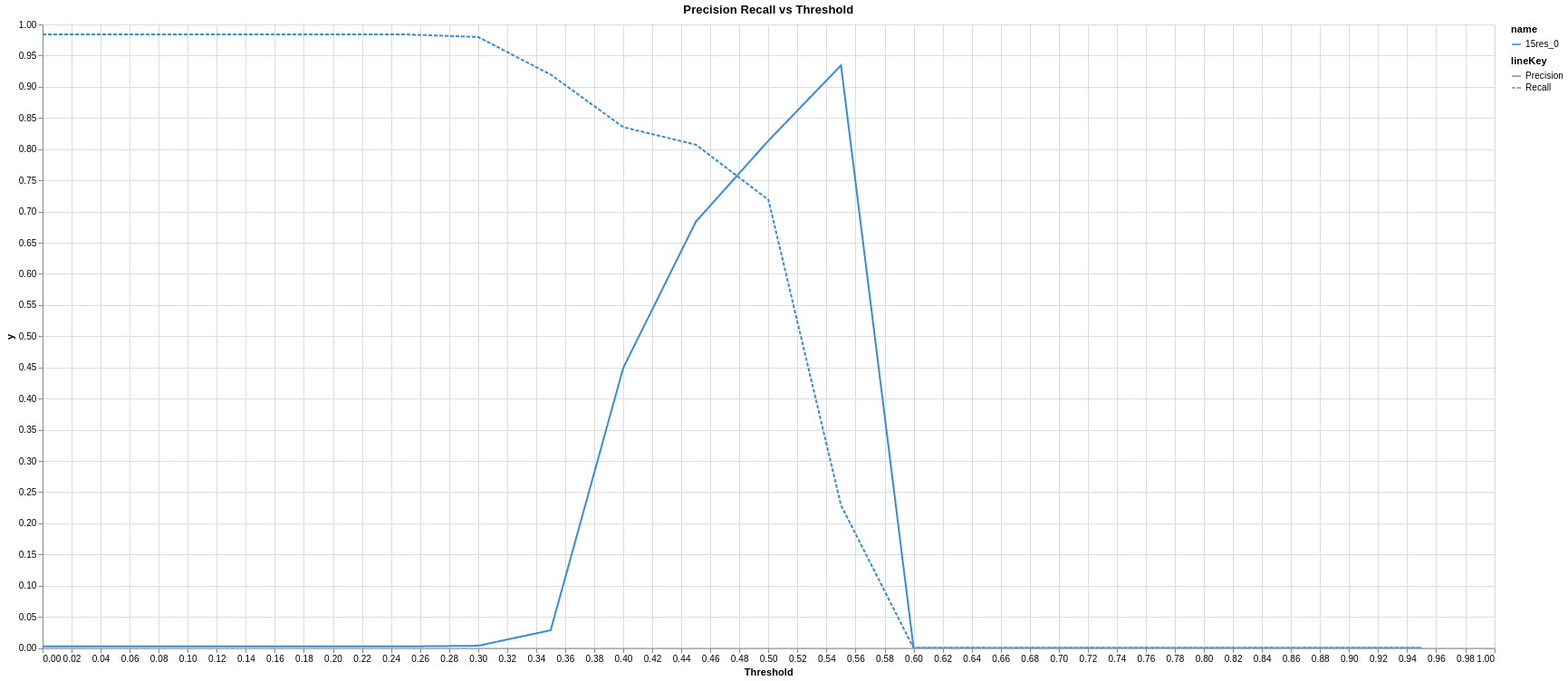}
    \caption{Precision and Recall as a function of $\tau$.}
    \label{fig:tau}
\end{figure*}

Figure~\ref{fig:tau} presents the relation between the value of $\tau$ and precision/recall on the 15res dataset. Such a plot can be constructed on a validation (or even training) set and used to guide the manual selection of $\tau$ hyperparameter. 
One heuristic for choosing $\tau$ is to start at the intersection of the precision and recall curves and then lower $\tau$ until the precision does not drop off abruptly (a knee point).
Of course, one could use standard hyperparameter selection methods, but we found this heuristic to be faster and more effective.

\section{Error analysis}\label{app:error}
To better understand how particular parts of our model influence the result, we measured precision/recall/F1 for several intermediate tasks: 1) the performance of an additional binary linear classifier trained to predict whether a span is valid based on the span representation $s_i$ (Span binary) , 2) the performance of extracting correct aspect-opinion pairs in the aspect-opinion pair construction layer (A-O pair layer), 3) the classification performance of the final classifier that assigns sentiment to the triples or filters them (Final 4-class), 4) the classification performance of the same classifier measured for the binary classification valid/invalid triple (Final binary).
The results are presented in Tab.~\ref{tabapp:error}.

\begin{table*}[]
\small\centering
\begin{tabular}{l|lll|lll|lll|lll}
\toprule
               & \multicolumn{3}{c}{14lap} & \multicolumn{3}{c}{14res} & \multicolumn{3}{c}{15res} & \multicolumn{3}{c}{16res} \\
     Dataset          & Prec.   & Rec.   & F1     & Prec.   & Rec.   & F1     & Prec.   & Rec.   & F1     & Prec.   & Rec.   & F1     \\\midrule
Span Binary    & 84.10   & 85.62  & 84.84  & 86.40   & 91.86  & 89.05  & 85.37   & 89.14  & 87.21  & 84.18   & 93.10  & 88.41  \\
A-O pair layer & 68.68   & 72.78  & 66.34  & 73.13   & 83.95  & 78.16  & 71.11   & 78.14  & 74.45  & 75.33   & 82.64  & 78.79  \\
Final 4-class  & 69.53   & 69.53  & 69.53  & 77.12   & 77.12  & 77.12  & 72.07   & 72.07  & 72.07  & 75.79   & 75.79  & 75.79  \\
Final binary   & 97.21   & 76.00  & 85.28  & 97.31   & 81.70  & 88.82  & 98.36   & 76.84  & 86.27  & 99.06   & 80.07  & 88.56  \\\midrule
Overall        & 67.58   & 62.48  & 64.90  & 76.43   & 75.70  & 76.06  & 72.91   & 71.34  & 72.10  & 76.27   & 76.12  & 76.19 \\
\bottomrule
\end{tabular}
\caption{The predictive performance of several parts of ASTE-Transformer on four benchmarks.}
\label{tabapp:error}
\end{table*}

\section{Characteristics of ASTE problems requiring dependency modeling}
\label{app:characteristics}
The following is an extended, but still non-exhaustive, list of properties unexplored by previous span-level ASTE approaches. All these properties have a common feature: they cannot be exploited due to the strong independence assumption made by span-level ASTE approaches when constructing the results. The proposed method, ASTE-Transformer, addresses this by relaxing this assumption and modelling the dependencies between the output triples. 
\begin{itemize}
    \item A phrase should be of the same type in all triples. For instance, if a given phrase is an aspect phrase then in other triples it can not be an opinion phrase.
    \item An opinion phrase assigned to multiple aspects, typically assign them the same sentiment polarity (see the 2$^{nd}$ example in Fig~\ref{fig:example}). 
    \item Two opinion phrases linked with a contrastive conjunction (like "but") and attached to one aspect phrase should have different sentiment polarities. A similar rule applies to opinions linked with correlative conjunctions ("and").
    \item Construction of a triple with a given aspect/opinion phrase should invalidate triples with overlapping phrases. 
    For example, if the model constructed the correct triple with "extremely pricy" when processing the second sentence in Fig.~\ref{fig:example}, then all candidate triples with "pricy" should be discarded\footnote{Some authors mention using additional post-processing to remove triples with overlapping phrases using heuristics. Still, this is external to the model and the model is not able to learn to exploit this dependency from the data.}. 
    The same is true for multi-word aspect phrases.
    \item Although one-to-many relations between aspect and opinion phrases are possible, the general probabilistic property is that constructing an increasing number of triples with a given phrase should be less and less likely.
    \item 
    A potential aspect phrase should only be extracted if there is an opinion phrase associated with it. For example, consider the aspect word "room" in "The room was fine..." and "I was given a single room". In the first sentence this word should be extracted because it is part of a triple (see Fig~\ref{fig:example}), but in the second sentence there is no opinion phrase attached to it.
\end{itemize}
\end{document}